\documentclass[11pt]{article}
\usepackage{acl}

\usepackage[T1]{fontenc}
\usepackage[utf8]{inputenc}
\usepackage{times}
\usepackage{latexsym}
\usepackage{microtype}
\usepackage{inconsolata}
\usepackage{amsmath}
\usepackage{amssymb}
\usepackage{booktabs}
\usepackage{enumitem}
\usepackage{graphicx}
\usepackage{tabularx}
\usepackage{array}
\usepackage{verbatim}
\usepackage{subcaption}

\usepackage{cuted}
\usepackage{caption}
\usepackage{multirow}
\usepackage[table]{xcolor}

\usepackage{algorithm}
\usepackage{algpseudocode}

\usepackage[most]{tcolorbox}
\usepackage{listings}
\usepackage{xcolor}

\lstdefinestyle{promptstyle}{
  basicstyle=\ttfamily\scriptsize,
  breaklines=true,
  columns=fullflexible,
  keepspaces=true,
  showstringspaces=false,
  frame=none,
  aboveskip=0pt,
  belowskip=0pt
}

\newtcblisting{promptbox}{
  enhanced,
  breakable,
  colback=white,
  colframe=white,
  boxrule=0pt,
  arc=0mm,
  boxsep=0pt,
  left=0pt,
  right=0pt,
  top=0pt,
  bottom=0pt,
  listing only,
  listing options={
    basicstyle=\ttfamily\footnotesize,
    breaklines=true,
    breakindent=0pt,
    postbreak=\mbox{},
    columns=fullflexible,
    keepspaces=true,
    showstringspaces=false,
    xleftmargin=0pt,
    xrightmargin=0pt,
    aboveskip=0pt,
    belowskip=0pt
  }
}

\title{HyperProve: Answer-Guided Hypergraph Expansion\\for Multi-Hop Question Answering
}

\author{
  \textbf{An Nguyen Phu}\textsuperscript{1}\thanks{Equal contribution},
  \textbf{Dung Nguyen Quang}\textsuperscript{1}\footnotemark[1],
  \textbf{Luu Hieu An}\textsuperscript{1}, \\
  \textbf{Linh Ngo Van}\textsuperscript{1}\thanks{Corresponding author: \href{mailto:linhnv@soict.hust.edu.vn}{linhnv@soict.hust.edu.vn}},
  \textbf{Trung Le}\textsuperscript{2},
  \textbf{Thien Huu Nguyen}\textsuperscript{3} \\[0.45em]
  {\normalfont\textsuperscript{1}Hanoi University of Science and Technology} \\
  {\normalfont\textsuperscript{2}Monash University,\quad
  \textsuperscript{3}University of Oregon} \\[0.35em]
}

\begin{document}
\maketitle

\begin{abstract}
Multi-hop question answering often fails when retrieval treats evidence as isolated matches to the original question, since the facts needed to answer a complex question are usually connected through intermediate entities, relations, and constraints. We propose \textbf{HyperProve}, a retrieval-augmented QA framework that addresses this challenge by coupling question decomposition with answer-conditioned expansion over a hypergraph of atomic facts. HyperProve does not use atomic facts, hypergraphs, or iterative retrieval in isolation; instead, it carries intermediate answers and supporting hyperedges as retrieval state, then uses that state to bias the next local hypergraph expansion. This design enables HyperProve to construct coherent evidence chains for final answer generation while making the retrieval process stateful and fact-centered. Across multi-hop QA benchmarks, HyperProve achieves the best overall performance in our evaluation, outperforming the strongest baselines by an average relative improvement of 6.2\% in answer accuracy and 4.9\% in F1.

\end{abstract}

\section{Introduction}

Retrieval-augmented generation~\citep{lewis2020retrieval} has become a standard way to ground language models in external evidence, but many retrieval pipelines still treat question answering as a one-shot relevance problem: given a question, retrieve passages that are independently similar to it, then ask a reader to generate the answer. This setup is better suited to questions whose answer-bearing evidence is directly aligned with the query. In multi-hop question answering, however, the crucial evidence is often distributed across facts connected through intermediate entities, relations, temporal constraints, or other bridge conditions~\citep{trivedi2022musique,yang2018hotpotqa,ho2020constructing}. In such cases, the next useful fact may not be similar to the original question at all; it becomes relevant only after an earlier fact has established the right bridge or connection.

\begin{figure}[t]
    \centering
    \includegraphics[width=\columnwidth]{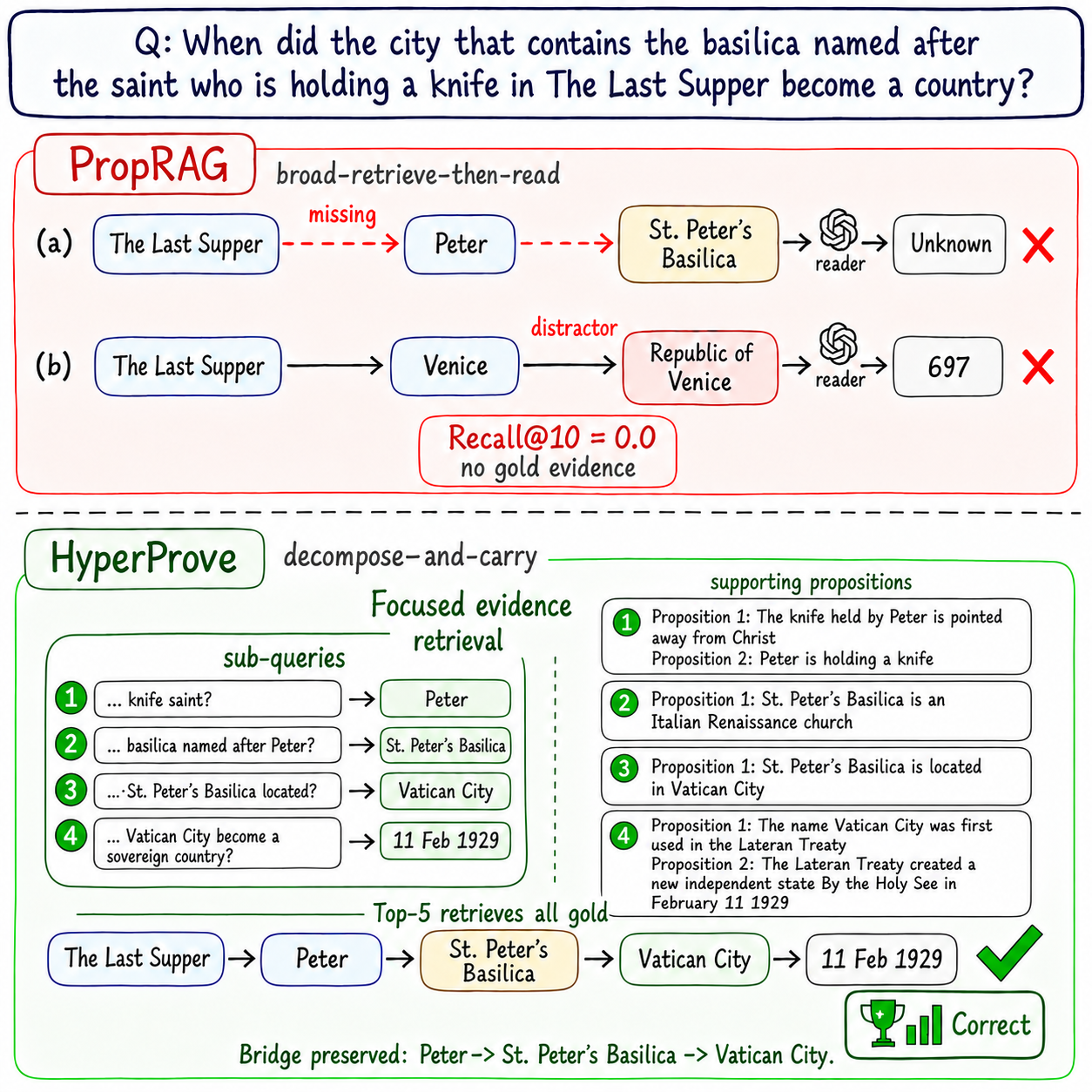}
    \caption{Case study of a 4-hop question: PropRAG fails to connect \textit{Last Supper} to \textit{Peter} and drifts to the distractor chain \textit{Venice} $\rightarrow$ \textit{Republic of Venice}; HyperProve succeeds because atomic-fact hyperedges explicitly preserve the bridge entities and correct path.}
    \label{fig:hyperprove-proprag-case}
\end{figure}

Prior work attacks this problem through graph structure and iterative retrieval, but its retrieval control differs from ours. HyperGraphRAG represents $n$-ary facts as hyperedges, yet matches nodes against the original question and applies a one-shot bipartite incidence expansion~\citep{luo-etal-2025-hypergraphrag}. HGRAG instead represents whole passages as hyperedges and combines entity- and passage-level relevance through fixed-query hypergraph diffusion~\citep{wang2026cross}. PropRAG preserves context in propositions and performs efficient, LLM-free beam search over proposition paths scored against the unchanged question~\citep{wang2025proprag}. These methods improve structured evidence discovery, but graph adjacency or original-query relevance, rather than a validated intermediate answer and its support, determines how retrieval proceeds. As Figure~\ref{fig:hyperprove-proprag-case} illustrates, a fixed-query proposition path can retain a globally similar distractor after the unresolved bridge has changed.

Iterative retrieval methods decompose questions, rewrite queries, or interleave retrieval with reasoning~\citep{trivedi2023ircot,wang2025sgfsm,ye2025qdream,zhu2025chainrag}. SG-FSM, for example, dynamically advances among sub-questions using textual state, while IRCoT and ChainRAG use intermediate reasoning or rewritten queries to guide later retrieval. Such systems adapt the query across hops, but they do not convert the resolved answer together with its supporting facts into a structural retrieval frontier. A later hop can therefore restart from surface similarity, lose the bridge evidence, or follow a plausible distractor entity.

We propose \textbf{HyperProve}, a retrieval-augmented QA framework that makes retrieval state explicit at the level of source-grounded atomic-fact hyperedges. Its key coupling is \emph{resolved answer + supporting atomic facts $\rightarrow$ next-hop structural frontier}. HyperProve decomposes a question into dependent sub-queries, rewrites each unresolved relation with validated bridge answers, combines the carried frontier with fresh dense seeds, and performs bounded query-aware fact-to-fact expansion. Unlike one-shot incidence expansion, fixed-query diffusion, or textual answer propagation, this mechanism changes the next retrieval distribution using both the resolved answer and the evidence that supports it. In Figure~\ref{fig:hyperprove-proprag-case}, the resulting frontier preserves the hidden bridge evidence and keeps the chain on the correct path.

Our main contributions are summarized as follows:
\begin{itemize}
    \item We propose \textbf{HyperProve}, an answer-conditioned hyperedge retrieval formulation that unifies graph-based and iterative RAG: each intermediate answer is carried with its supporting atomic-fact hyperedges as the structural frontier for subsequent hops.

    \item We design a query-biased local hypergraph expansion algorithm that retrieves over atomic-fact hyperedges, combining structural connectivity among evidence units with semantic alignment to each rewritten sub-query.

    \item Under a controlled retriever, reader, corpus, and evidence budget, we show that HyperProve improves multi-hop QA across three benchmarks over dense, structured, and iterative retrieval baselines.
\end{itemize}

\begin{figure*}[t]
    \centering
    \includegraphics[width=1.0\linewidth]{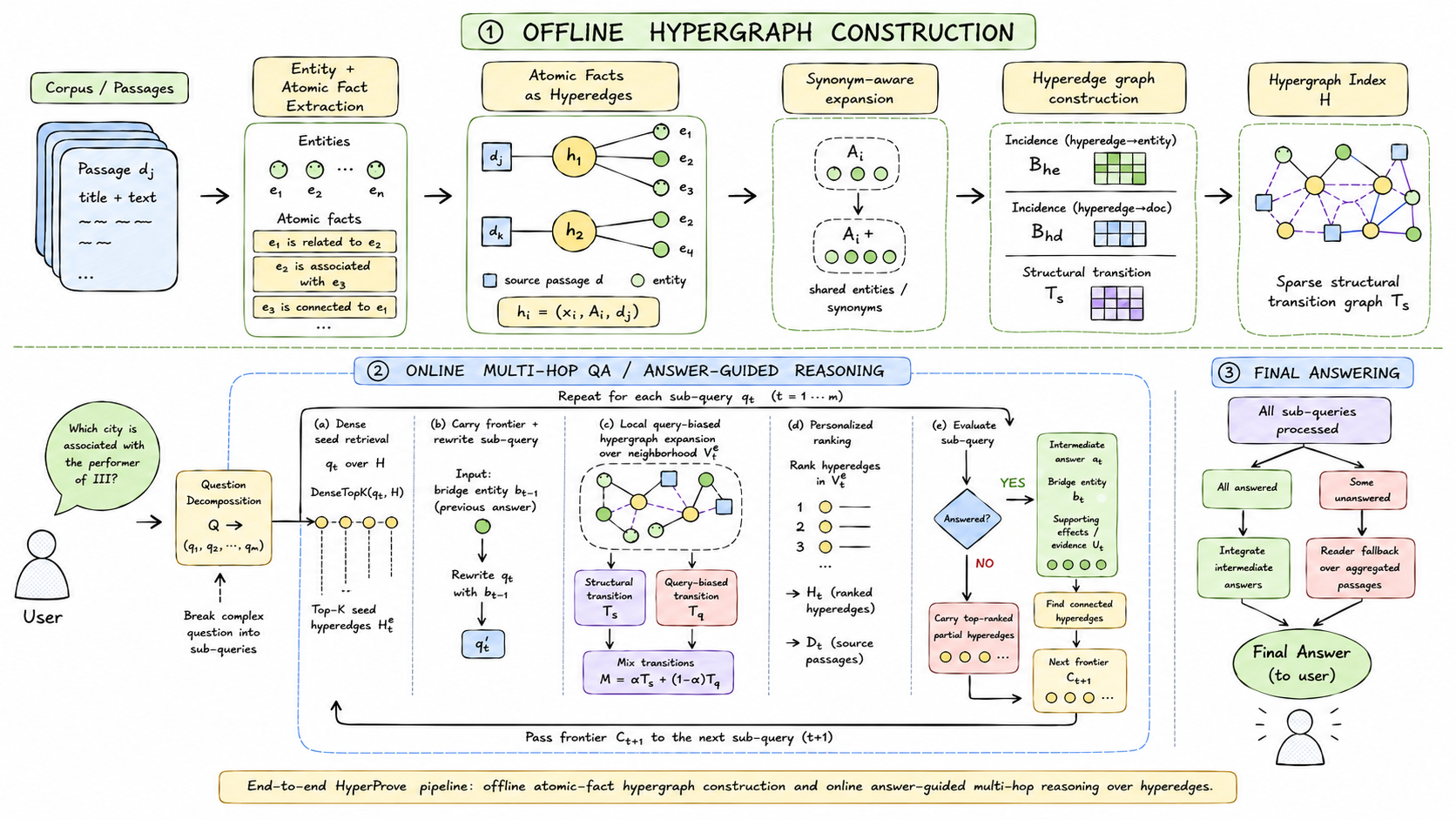}
    \caption{Overall HyperProve pipeline. The system builds an offline synonym-aware atomic-fact hypergraph from corpus passages, then performs online answer-guided multi-hop reasoning by decomposing a question into sub-queries, carrying intermediate answers as bridge entities, expanding over local hypergraph neighborhoods, and integrating the resulting evidence chain into a final answer.}
    \label{fig:overall_pipeline}
\end{figure*}

\section{Related Work}

\paragraph{Graph-based RAG.}
Graph-based RAG improves over dense retrieval by modeling relationships among passages, entities, or atomic facts instead of scoring passages independently. HippoRAG2 builds graph memory with Personalized PageRank~\citep{gutierrez-etal-2025-hipporag2}; PropRAG searches proposition paths to retain fine-grained context~\citep{wang2025proprag}; HyperGraphRAG represents $n$-ary facts as hyperedges~\citep{luo-etal-2025-hypergraphrag}; and HGRAG uses an entity--passage hypergraph for multi-hop QA~\citep{wang2026cross}. The same structural view extends beyond English open-domain QA: MaGiX builds a multi-granular cross-lingual graph with cross-synonym edges for English--Vietnamese RAG~\citep{hieu2025magix}, and MemORAI organizes conversational memory in a provenance-aware multi-relational graph traversed by Dynamic Weighted PageRank~\citep{pham2026memorai}. Cross-lingual retrievers trained with symmetric objectives further improve the underlying matching quality~\citep{nguyen2025improving}. However, they remain suboptimal for multi-hop QA because their retrieval stays conditioned on the original question, whereas HyperProve carries each intermediate answer and its supporting hyperedges across sub-queries to guide query-biased local expansion over atomic facts.

\paragraph{Iterative Retrieval.}
Iterative methods plan, retrieve, and refine evidence across multiple steps. IRCoT interleaves retrieval with chain-of-thought reasoning~\citep{trivedi2023ircot}; SG-FSM advances through dynamically selected sub-questions using a textual finite-state process~\citep{wang2025sgfsm}; T2RAG reasons over triples~\citep{gong2025t2rag}; HopRAG performs retrieve--reason--prune steps~\citep{liu2025hoprag}; and ChainRAG progressively rewrites queries to mitigate lost-in-retrieval errors~\citep{zhu2025chainrag}. KiRAG and UniRAG further adapt retrieval through knowledge-driven iteration and decomposition~\citep{fang2025kirag,kim2025unirag}. HyperProve shares their hop-adaptive view, but changes the type of state: the validated answer and its supporting atomic facts define the next structural frontier, while the rewritten sub-query biases transitions within that bounded neighborhood. This preserves evidence provenance across hops rather than carrying only text into a new retrieval call.

\section{Methodology}

Let $\mathcal{D}=\{d_j\}_{j=1}^{M_d}$ be document nodes and let $p_j=\operatorname{text}(d_j)$ denote the passage stored at node $d_j$. Given question $Q$, our goal is to retrieve a compact set of connected evidence and generate answer $a$. We view a multi-hop question as a sequence of dependent sub-queries,
\begin{equation*}
Q \rightarrow (q_1,q_2,\ldots,q_m),
\end{equation*}
where each $q_t$ resolves at least one bridge entity, bridge relation, or final target attribute. The answer or evidence from $q_{t-1}$ conditions the interpretation and retrieval frontier of $q_t$. HyperProve operationalizes this view by constructing a hyperedge-level knowledge structure and performing all intermediate reasoning directly over atomic-fact hyperedges. Figure~\ref{fig:overall_pipeline} summarizes the resulting offline indexing and online answer-guided reasoning pipeline.

\subsection{Atomic Facts as Hyperedges}

As the first offline stage in Figure~\ref{fig:overall_pipeline}, HyperProve extracts entity nodes from each passage $p_j$,
\[
\mathcal E_j\subseteq\mathcal V=\{e_a\}_{a=1}^{M_e},
\]
where $m_a=\operatorname{name}(e_a)$ is the textual mention associated with entity node $e_a$. The extracted nodes include named entities, dates, salient objects, and generic concepts that participate in meaningful relations. The passage is then decomposed into atomic facts. Let $\mathcal H=\{h_i\}_{i=1}^{N}$ be the resulting hyperedges and let $\sigma(i)=j$ map hyperedge $h_i$ to its source document. We represent each atomic fact as
\[
h_i=(x_i,\mathcal A_i,d_{\sigma(i)}),
\]
where $x_i=\operatorname{text}(h_i)$ is a self-contained factual statement and $\mathcal A_i\subseteq\mathcal E_{\sigma(i)}$ is the set of participating entity nodes. Structurally, its incidences are
\[
h_i\sim\mathcal A_i\cup\{d_{\sigma(i)}\}.
\]

This modeling choice addresses the limitations of binary triples. A triple represents a pairwise relation $(s,r,o)$, but many facts involve more than two participants together with temporal, spatial, causal, or provenance qualifiers. Splitting such a fact into disconnected pairs can cause \emph{context collapse}. PropRAG similarly motivates context-rich propositions over triples~\citep{wang2025proprag}; HyperProve additionally makes each atomic fact a hyperedge that connects all participating entity nodes and its source document. Extraction prompts are provided in Appendix~\ref{app:prompt-templates}.

\subsection{Knowledge Hypergraph Construction}

The next offline stage turns the corpus into a synonym-aware hyperedge graph. Entity surface forms may differ across passages, so HyperProve embeds entity names $m_a$ and creates a synonym map using cosine similarity. If two name embeddings exceed threshold $\eta$ (0.8 in our experiments), their entity nodes are treated as neighboring mentions. The participant set is expanded as
\[
\mathcal A_i^+ = \mathcal A_i \cup \operatorname{Syn}(\mathcal A_i).
\]
The expanded set allows hyperedges to overlap even when passages use different surface forms for the same entity or concept.

We define hyperedge--entity incidence $B_{he}\in\{0,1\}^{N\times M_e}$ and hyperedge--document incidence $B_{hd}\in\{0,1\}^{N\times M_d}$:
\[
\begin{aligned}
B_{he}[i,a] &= \mathbf{1}[e_a\in\mathcal A_i^+],\\
B_{hd}[i,j] &= \mathbf{1}[\sigma(i)=j].
\end{aligned}
\]
The entity and document degree matrices are
\[
\begin{aligned}
D_e[a,a] &= \sum_i B_{he}[i,a],\\
D_d[j,j] &= \sum_i B_{hd}[i,j].
\end{aligned}
\]

The weighted structural adjacency between hyperedges is
\[
W_s=B_{he}D_e^{-1}B_{he}^{\top}+B_{hd}D_d^{-1}B_{hd}^{\top}.
\]
The first term connects hyperedges that share expanded entity nodes, while the second connects hyperedges from the same document. $D_e^{-1}$ down-weights common entities and $D_d^{-1}$ down-weights documents that yield many atomic facts. After removing self-loops, we row-normalize:
\[
T_s=\operatorname{RowNormalize}\!\left(W_s-\operatorname{diag}(W_s)\right).
\]
The resulting $T_s$ is a sparse transition matrix over atomic-fact hyperedges; row normalization is applied only after self-transitions are removed. Documents remain provenance and reader context, while retrieval transitions operate over $\mathcal H$.

\subsection{Query-biased Hyperedge Retrieval}

For each query or sub-query, HyperProve uses dense retrieval only to select entry hyperedges. These seeds define where graph search begins; evidence is then ranked by navigating connected atomic facts in the hyperedge graph.

Starting from the current seeds, HyperProve builds local subgraph $G_t^\star$ by random-walk expansion over $T_s$, retaining at most 600 hyperedges connected through entities, synonym-expanded overlap, or shared provenance. Let $T_{s,t}^\star$ be the row-normalized restriction of $T_s$ to $G_t^\star$. For candidate hyperedge $h_k$, define
\[
\begin{aligned}
c_{t,k}&=\cos\!\left(\phi(q'_t),\phi(x_k)\right),\\
\mathcal N_{t,\theta}(i)&=\{k\in\mathcal N_t(i):c_{t,k}>\theta\},\\
Z_{t,i}&=\sum_{\ell\in\mathcal N_{t,\theta}(i)}\exp(c_{t,\ell}/\tau).
\end{aligned}
\]
The query-biased transition is
\[
T_{q,t}^\star[i,k] =
\begin{cases}
\exp(c_{t,k}/\tau)/Z_{t,i}, & k\in\mathcal N_{t,\theta}(i),\\
0, & \text{otherwise},
\end{cases}
\]
where $\mathcal N_t(i)$ is the structural neighborhood of $h_i$ in $G_t^\star$. Each row therefore assigns mass only to structurally connected hyperedges whose similarity to the current rewritten sub-query exceeds $\theta$. If no candidate passes the gate, the row is zero and retrieval relies on the structural transition.

The final transition matrix mixes structural connectivity and query relevance:
\[
M_t = \alpha T_{s,t}^\star + (1-\alpha)T_{q,t}^\star.
\]
Here, $\alpha$ controls how much the walk follows the hypergraph structure versus the current query semantics. Personalized ranking is then performed over this local hyperedge graph:
\[
\boldsymbol{\pi}^{(r+1)}_t=(1-\gamma)u_t^\star+\gamma M_t^\top \boldsymbol{\pi}^{(r)}_t,
\]
where $r$ indexes power iteration and $u_t^\star$ resets to the current seeds. We use $\tau=1$, $\theta=0.4$, $\alpha=0.5$, and $\gamma=0.85$ throughout. The resulting $\boldsymbol\pi_t$ ranks atomic-fact hyperedges; documents are recovered only after hyperedge ranking through the source map $\sigma$.

\subsection{Answer-Guided Hyperedge Reasoning}

HyperProve decomposes a question into dependent sub-queries and carries evidence across them as a hyperedge frontier; Algorithm~\ref{alg:hyperprove_agentic} gives the procedure. Given $Q\rightarrow(q_1,\ldots,q_m)$, let $\mathcal C_1=\emptyset$ and, for $t>1$, let $\mathcal C_t\subseteq\mathcal H$ be the frontier produced after hop $t-1$. Each hop uses at most $K_c=10$ carried-frontier seeds and $K_d=10$ fresh dense seeds. Fresh seeds allow retrieval to escape a disconnected or incorrect local neighborhood. Let $\mathcal F_t$ denote the fresh seeds; then
\[
\begin{aligned}
    \mathcal F_t &= \operatorname{DenseTopK}(q'_t,\mathcal H,K_d),\\
    \mathcal S_t &=
    \begin{cases}
    \mathcal F_t, & t=1,\\
    \operatorname{TopK}(\mathcal C_t,K_c) \cup \mathcal F_t, & t>1.
    \end{cases}
\end{aligned}
\]
For the first hop, $q'_1=q_1$. For $t>1$, let $\mathcal R_{t-1}=\{(q_s,a_s):s<t,\ s_s=\textsc{Answered}\}$ be the resolved question--answer memory. Then $q'_t=\operatorname{Rewrite}(q_t,\mathcal R_{t-1})$, which replaces referring expressions with validated intermediate answers while preserving the unresolved relation. If $\mathcal R_{t-1}$ is empty, $q'_t=q_t$.

When a sub-query is answered, bridge entity $b_t$ and the entities in supporting hyperedges $\mathcal U_t$ define the next frontier. Let $\mathcal E_t=\operatorname{Entities}(\mathcal U_t)$; then
\[
\mathcal C_{t+1}
=
\operatorname{FindHyperedges}
\bigl(\{b_t\} \cup \mathcal E_t, \mathcal H\bigr).
\]
If the sub-query is unresolved, HyperProve carries the top-$K_c$ hyperedges under $\boldsymbol\pi_t$ as partial evidence.

\begin{algorithm}[t]
\scriptsize
\caption{HyperProve main algorithm.}
\label{alg:hyperprove_agentic}

\begin{algorithmic}[1]
\Require Hyperedge index $(\mathcal H,\mathcal V,\mathcal D,T_s)$, question $Q$, $K_c=K_d=10$
\Ensure Final answer $a$

\State $(q_1,\ldots,q_m) \gets \textsc{Decompose}(Q)$
\State $\mathcal M_H,\mathcal M_D,\mathcal C_1,\mathcal R \gets \emptyset$
\Statex \hspace{\algorithmicindent}\textit{$\mathcal M_H$: hyperedge memory; $\mathcal M_D$: document memory}

\For{$t=1$ to $m$}
    \If{$t=1$}
        \State $q'_t\gets q_t$; $\mathcal S_t \gets \textsc{DenseTopK}(q'_t,\mathcal H,K_d)$
    \Else
        \State $q'_t \gets \textsc{Rewrite}(q_t,\mathcal R)$
        \State $\mathcal S_t \gets \textsc{TopK}(\mathcal C_t,K_c) \cup \textsc{DenseTopK}(q'_t,\mathcal H,K_d)$
    \EndIf

    \State retrieve $\mathcal H_t$, scores $\boldsymbol\pi_t$, passages $\mathcal P_t$ from $\mathcal S_t$
    \State $\mathcal M_H \gets \mathcal M_H \cup \{(\mathcal H_t,\boldsymbol\pi_t)\}$
    \State $\mathcal M_D \gets \textsc{MaxMerge}(\mathcal M_D,\mathcal P_t)$
    \State $(s_t,a_t,\mathcal U_t) \gets \textsc{Evaluate}(q_t,\mathcal H_t,\mathcal P_t)$

    \If{$s_t=\textsc{Answered}$}
        \State $\mathcal R \gets \mathcal R \cup \{(q_t,a_t)\}$
        \State $b_t \gets \textsc{BridgeEntity}(a_t)$; $\mathcal E_t \gets \textsc{Entities}(\mathcal U_t)$
        \State $\mathcal C_{t+1} \gets \textsc{FindHyperedges}(\{b_t\} \cup \mathcal E_t,\mathcal H)$
    \Else
        \State $\mathcal C_{t+1} \gets \textsc{TopK}(\mathcal H_t,\boldsymbol\pi_t,K_c)$
    \EndIf
\EndFor

\If{all sub-queries are answered}
    \State $a \gets \textsc{IntegrateAnswers}(Q,\mathcal R)$
\Else
    \State $\mathcal P_{10} \gets \textsc{TopKPassages}(\mathcal M_H,\mathcal M_D,10)$
    \State $a \gets \textsc{ReaderFallback}(Q,\mathcal P_{10},\mathcal M_H)$
\EndIf

\State \Return $a$
\end{algorithmic}
\end{algorithm}

\paragraph{Reader fallback.}
If every sub-query is answered, $\operatorname{IntegrateAnswers}(Q,\mathcal R)$ produces the final answer. Otherwise, $\operatorname{ReaderFallback}$ max-merges evidence across hops and invokes the same reader LLM once with the original question. Given scored hyperedge memory $\mathcal M_H=\{(\mathcal H_t,\boldsymbol\pi_t)\}_{t=1}^{m}$, document $d_j$ receives score
\[
    s(d_j)
    =
    \max_{\substack{t\in[m],\,h_i\in\mathcal H_t\\
    \sigma(i)=j}}
    \pi_t(h_i).
\]
Let $\mathcal P_{10}$ contain the ten documents with highest $s(d_j)$. The fallback reader receives $Q$, their passage texts $p_j$, and the corresponding ranked atomic facts from $\mathcal M_H$. Appendix~\ref{app:prompt-qa-fallback} and \ref{app:fallback-analysis} provide further details.

\section{Experiments}

\begin{table*}[t]
\centering
\scriptsize
\newcommand{\score}[2]{\mbox{#1{\tiny$\,\pm\,$#2}}}
\setlength{\tabcolsep}{3.8pt}
\renewcommand{\arraystretch}{1.08}
\caption{Mean $\pm$ standard deviation over three runs on multi-hop QA benchmarks. We report LLM-as-a-judge answer accuracy, Exact Match (EM), and token-level F1. Best means are in \textbf{bold}; second-best means are \underline{underlined}.}
\label{tab:main_results}
\resizebox{0.92\textwidth}{!}{
\begin{tabular}{lccccccccc}
\toprule
& \multicolumn{3}{c}{\textbf{MuSiQue}} 
& \multicolumn{3}{c}{\textbf{HotpotQA}} 
& \multicolumn{3}{c}{\textbf{2Wiki}} \\
\cmidrule(lr){2-4} \cmidrule(lr){5-7} \cmidrule(lr){8-10}
\textbf{Method} 
& Acc. & EM & F1
& Acc. & EM & F1
& Acc. & EM & F1 \\
\midrule

\rowcolor{gray!15}
\multicolumn{10}{c}{\textbf{\textit{Dense Retrieval}}} \\
Contriever
& \score{47.80}{0.28} & \score{22.30}{0.30} & \score{32.10}{0.26}
& \score{55.55}{0.21} & \score{44.40}{0.24} & \score{57.50}{0.21}
& \score{69.60}{0.20} & \score{38.50}{0.24} & \score{43.70}{0.21} \\

BGE-M3 
& \score{51.50}{0.25} & \score{27.80}{0.27} & \score{39.70}{0.24}
& \score{68.12}{0.19} & \score{47.80}{0.22} & \score{61.80}{0.19}
& \score{72.18}{0.18} & \score{48.20}{0.22} & \score{54.50}{0.19} \\

\rowcolor{gray!15}
\multicolumn{10}{c}{\textbf{\textit{Structured Retrieval}}} \\
HippoRAG2 
& \score{52.00}{0.36} & \score{37.20}{0.40} & \score{48.60}{0.34}
& \score{81.70}{0.29} & \score{62.70}{0.33} & \underline{\score{75.50}{0.29}}
& \score{86.40}{0.28} & \score{65.00}{0.34} & \score{71.00}{0.29} \\

PropRAG 
& \score{50.90}{0.30} & \score{29.30}{0.33} & \score{46.73}{0.29}
& \underline{\score{85.80}{0.24}} & \score{57.80}{0.28} & \score{73.79}{0.24}
& \score{77.50}{0.23} & \score{58.00}{0.29} & \score{68.78}{0.24} \\

HyperGraphRAG 
& \score{55.13}{0.43} & \score{42.20}{0.47} & \score{53.80}{0.40}
& \score{77.89}{0.34} & \underline{\score{63.90}{0.39}} & \score{73.80}{0.33}
& \score{77.72}{0.33} & \underline{\score{68.30}{0.40}} & \score{71.30}{0.33} \\

HGRAG
& \score{58.60}{0.39} & \score{41.19}{0.43} & \underline{\score{55.04}{0.37}}
& \score{80.10}{0.31} & \score{54.90}{0.36} & \score{69.43}{0.30}
& \score{85.00}{0.30} & \score{64.10}{0.37} & \score{74.10}{0.31} \\

\rowcolor{gray!15}
\multicolumn{10}{c}{\textbf{\textit{Iterative Retrieval}}} \\
IRCoT  
& \score{56.67}{0.64} & \score{34.10}{0.72} & \score{47.60}{0.59}
& \score{85.23}{0.49} & \score{55.70}{0.56} & \score{71.20}{0.47}
& \score{83.65}{0.47} & \score{60.70}{0.57} & \score{74.30}{0.47} \\

T2RAG
& \score{54.67}{0.47} & \score{34.30}{0.52} & \score{45.60}{0.44}
& \score{81.33}{0.36} & \score{54.20}{0.42} & \score{67.30}{0.35}
& \score{83.75}{0.35} & \score{66.70}{0.43} & \score{74.40}{0.36} \\

HopRAG 
& \underline{\score{67.18}{0.72}} & \underline{\score{42.40}{0.80}} & \score{54.90}{0.66}
& \score{85.51}{0.55} & \score{61.10}{0.63} & \score{68.26}{0.52}
& \underline{\score{86.66}{0.53}} & \score{62.00}{0.64} & \underline{\score{76.06}{0.52}}\\

ChainRAG 
& \score{60.01}{0.66} & \score{41.20}{0.74} & \score{54.15}{0.61}
& \score{85.46}{0.50} & \score{60.70}{0.58} & \score{74.66}{0.48}
& \score{82.48}{0.49} & \score{63.30}{0.60} & \score{74.39}{0.49} \\

\midrule
\rowcolor{gray!20}
\textbf{HyperProve} 
& \textbf{\score{71.60}{0.58}} & \textbf{\score{42.83}{0.64}} & \textbf{\score{59.23}{0.54}}
& \textbf{\score{91.30}{0.43}} & \textbf{\score{63.92}{0.49}} & \textbf{\score{76.64}{0.41}}
& \textbf{\score{91.40}{0.42}} & \textbf{\score{69.70}{0.51}} & \textbf{\score{80.01}{0.43}} \\

\bottomrule
\end{tabular}
}
\vspace{2pt}

\end{table*}

\subsection{Experimental Settings}

\paragraph{Datasets and Metrics.} We evaluate HyperProve on three open-domain multi-hop question answering benchmarks: MuSiQue~\cite{trivedi2022musique}, HotpotQA~\cite{yang2018hotpotqa}, and 2WikiMultiHopQA~\cite{ho2020constructing}. We report three answer-quality metrics: LLM-as-a-judge answer accuracy, Exact Match (EM), and token-level F1. 

\paragraph{Baselines.}
We compare HyperProve against three families of retrieval-augmented baselines. First, the dense retrieval family includes Contriever~\cite{izacard2022contriever} and BGE-M3~\cite{chen2024bgem3}. Second, structured retrieval includes graph- and structure-based RAG methods: HippoRAG2~\cite{gutierrez-etal-2025-hipporag2}, PropRAG~\cite{wang2025proprag}, HyperGraphRAG~\cite{luo-etal-2025-hypergraphrag}, and HGRAG~\cite{wang2026cross}. Third, the iterative retrieval family includes IRCoT~\cite{trivedi2023ircot}, T2RAG~\cite{gong2025t2rag}, HopRAG~\cite{liu2025hoprag}, and ChainRAG~\cite{zhu2025chainrag}.

\paragraph{Implementation Details.}
For all methods in the main experiments and analysis, we use \textsc{BGE-M3} as the retriever and \textsc{GPT-OSS-20B} as the backbone LLM. Apart from this shared model setup, each baseline follows its original implementation, including its hyperparameter configurations. Table~\ref{tab:main_results} reports the mean and standard deviation over three runs under this controlled setup.

\subsection{Main Results}

\paragraph{Dense retrieval lacks evidence state for multi-hop QA.}
Although BGE-M3 improves over Contriever, it remains far behind HyperProve: on MuSiQue, 51.50 accuracy and 39.70 F1 versus 71.60 and 59.23; on 2Wiki, 72.18 and 54.50 versus 91.40 and 80.01. Even on HotpotQA, where dense retrieval is stronger, BGE-M3 is still 23.18 accuracy points lower. These gaps show that independent dense matching can retrieve locally relevant passages, but lacks persistent evidence state for later hops once a bridge entity or constraint has been resolved.

\paragraph{Static structure and iterative retrieval are complementary but incomplete.} The strongest baseline alternates between Structured Retrieval and Iterative Retrieval methods: HopRAG leads accuracy on MuSiQue with 67.18 and 2Wiki with 86.66, while PropRAG leads HotpotQA accuracy with 85.80; HGRAG gives the strongest MuSiQue F1 with 55.04, and HyperGraphRAG gives the strongest baseline EM on HotpotQA with 63.90 and 2Wiki with 68.30. This uneven pattern reflects the limits of both families. Static structure helps connect evidence but uses a mostly fixed frontier that is not conditioned on intermediate answers, while iterative retrieval adapts the query across hops but can drift from the evidence that justified the previous hop.

\paragraph{HyperProve achieves the best evaluated performance.}
HyperProve has the highest mean on every reported metric across all three benchmarks. It improves answer accuracy over HopRAG by 4.42 points on MuSiQue and 4.74 points on 2Wiki; on HotpotQA, it improves over PropRAG by 5.50 accuracy points and over HippoRAG2 by 1.14 F1 points. The broad accuracy and F1 gains are larger than the observed run-level variation. We nevertheless avoid over-interpreting two narrow EM margins: $42.83$ versus $42.40$ on MuSiQue and $63.92$ versus $63.90$ on HotpotQA are within run-level variation and do not establish statistically resolved improvements. Overall, the results support using intermediate answers to define a structural frontier over atomic-fact hyperedges in addition to rewriting later queries.

\section{Analysis}
\label{sec:analysis}

We analyze HyperProve from four perspectives: component ablations, evidence-chain faithfulness, efficiency, and qualitative behavior. 
Together, these analyses examine which mechanisms drive multi-hop performance, whether it accumulates connected evidence across hops, whether HyperProve improves accuracy with reasonable cost, and how it handles hidden bridge entities and multi-entity reasoning.

\subsection{Retrieval Component Breakdown}
\label{sec:analysis_ablation}

\begin{table}[t]
\centering
\small
\setlength{\tabcolsep}{4pt}
\begin{tabular}{lcccc}
\toprule
\textbf{Variant} & \textbf{Overall} & \textbf{2-hop} & \textbf{3-hop} & \textbf{4-hop} \\
\midrule
HyperProve & \textbf{71.60} & \textbf{78.00} & \textbf{72.96} & \textbf{49.40} \\
w/o bridge entity & 64.50 & 69.11 & 67.41 & 44.58 \\
w/o fresh dense seed & 61.30 & 71.24 & 59.18 & 34.34 \\
w/o decomposition & 51.28 & 58.11 & 49.05 & 33.73 \\
w/o $T_s$ & 62.13 & 67.02 & 64.65 & 42.06 \\
w/o query transition & 58.17 & 65.20 & 58.60 & 35.40 \\
\bottomrule
\end{tabular}
\caption{Accuracy breakdown of HyperProve ablations on MuSiQue by reasoning depth. Accuracy is computed using judge verdicts over 1000 questions: 518 2-hop, 316 3-hop, and 166 4-hop examples.}
\label{tab:musique_ablation_by_hop}
\end{table}

The answer-guided ablations in Table~\ref{tab:musique_ablation_by_hop} show that HyperProve relies on both bridge propagation and fresh query-specific entry points. Removing bridge entities lowers overall accuracy from 71.60 to 64.50, with a substantial drop on 4-hop questions from 49.40 to 44.58, suggesting that intermediate answers provide cross-hop links that dense similarity alone may miss. Removing fresh dense seeds is also especially harmful for deeper reasoning: although 2-hop accuracy remains comparable, 4-hop accuracy falls to 34.34, indicating that the carried frontier alone can trap retrieval inside the current hyperedge neighborhood. The largest degradation comes from removing decomposition, which reduces overall accuracy to 51.28 and confirms that complex questions are better solved through dependent sub-queries than through a single retrieval step.

The transition-component ablations show that both $T_s$ and $T_{q,t}^{\star}$ are necessary for effective hyperedge reasoning. Removing the structural transition $T_s$ produces a pattern similar to removing bridge entities, but with a larger overall drop to 62.13, indicating that structural connectivity is crucial for maintaining valid evidence paths across hops. Removing the query-conditioned transition $T_{q,t}^{\star}$ causes an even sharper decline to 58.17 overall and 35.40 on 4-hop questions. This shows that structure alone is not enough: without query-aware transition, HyperProve no longer performs targeted graph reasoning and may expand along connected but irrelevant hyperedges.

\subsection{Evidence Chain Faithfulness}
\label{sec:analysis_faithfulness}

\begin{table}[t]
\centering
\small
\resizebox{\linewidth}{!}{
\begin{tabular}{lcccc}
\toprule
\textbf{Method} & \multicolumn{3}{c}{\textbf{Recall@10}} & \textbf{Avg Gain} \\
\cmidrule(lr){2-4}
& \textbf{MuSiQue} & \textbf{HotpotQA} & \textbf{2Wiki} & \textbf{After Hop 1} \\
\midrule
HyperProve & 82.31 & 96.20 & 98.15 & +41.24 \\
ChainRAG & 67.42 & 89.45 & 75.85 & +26.35 \\
HopRAG & 66.63 & 82.82 & 84.17 & +25.98 \\
\bottomrule
\end{tabular}
}
\caption{
Evidence chain faithfulness across multi-hop benchmarks.
Dataset columns report Recall@10, computed as the average fraction of gold passages retrieved in the top 10 results. Avg Gain after Hop 1 measures the average increase in cumulative gold-passage coverage after the first hop.
}
\label{tab:evidence_carry_effectiveness}
\end{table}

Table~\ref{tab:evidence_carry_effectiveness} evaluates evidence quality under the same top-10 retrieval budget across MuSiQue, HotpotQA, and 2Wiki, rather than only reporting answer accuracy. HyperProve achieves the highest Recall@10 on all three benchmarks, with 82.31 on MuSiQue, 96.20 on HotpotQA, and 98.15 on 2Wiki. These results improve over the strongest iterative baselines by 14.89, 6.75, and 13.98 points, respectively. HyperProve also obtains the largest average gain after the first hop, +41.24 versus +26.35 for ChainRAG and +25.98 for HopRAG. Thus, additional hops are not simply adding more context: they recover missing gold passages while preserving a stronger final support set than both iterative baselines.

\begin{figure}[t]
\centering
\includegraphics[width=\linewidth]{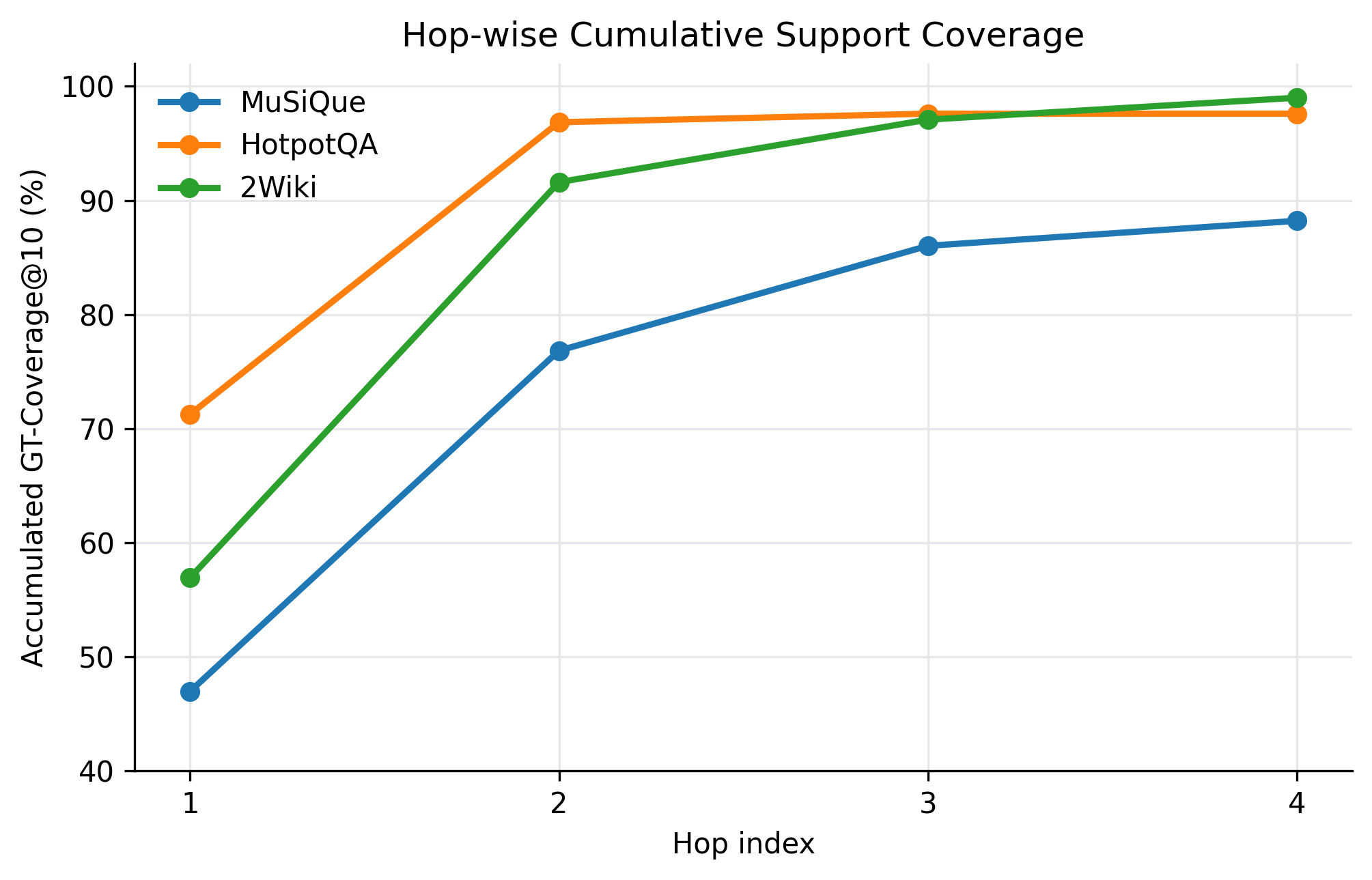}
\caption{
Hop-wise cumulative support coverage.
Cumulative gold-passage coverage increases across hops, showing that later sub-queries progressively recover missing support evidence.
}
\label{fig:hopwise_support_coverage}
\end{figure}

\begin{table*}[t]
\centering
\small
\renewcommand{\arraystretch}{1.15}
\begin{tabularx}{\textwidth}{l c X l}
\toprule
\textbf{System} & \textbf{Top-10 Coverage} & \textbf{Retrieval / Reasoning Behavior} & \textbf{Final Answer} \\
\midrule
HyperProve 
& 4 / 4 
& Separately resolves \textit{Pathology} $\rightarrow$ \textit{San Diego} and \textit{Still Swingin'} $\rightarrow$ \textit{Papa Roach} $\rightarrow$ \textit{California}, then applies the rank query to \textit{San Diego} under the \textit{California} urban-area constraint. 
& third-largest \\
\midrule
ChainRAG 
& 2 / 4 
& Retrieves evidence for \textit{Pathology} $\rightarrow$ \textit{San Diego} and \textit{Still Swingin'} $\rightarrow$ \textit{Papa Roach}, but misses the facts needed to bind \textit{Papa Roach} $\rightarrow$ \textit{Vacaville, California} and \textit{San Diego urban area} $\rightarrow$ \textit{third-largest in California}. Its reader is then pulled to a nearby city-rank answer. 
& 2nd \\
\bottomrule
\end{tabularx}
\caption{
Qualitative comparison on a multi-entity MuSiQue example.
HyperProve retrieves the complete evidence chain and preserves the urban-area constraint, while ChainRAG retrieves only partial evidence and confuses city population rank with urban-area rank.
}
\label{tab:case_multi_entity}
\end{table*}

Figure~\ref{fig:hopwise_support_coverage} further isolates the hop-wise behavior of HyperProve. Across all three benchmarks, each decomposed hop increases or maintains cumulative support coverage, with especially large gains after the first hop. Because the curve measures accumulated gold-passage coverage at Recall@10, these increases reflect newly recovered supporting evidence rather than longer reasoning traces or repeated retrievals.

\subsection{Efficiency Analysis}
\label{sec:analysis_efficiency}

\begin{table*}[t]
\centering
\begin{minipage}[t]{0.56\textwidth}
\centering
\textbf{(a) Offline construction}\par\smallskip
\scriptsize
\setlength{\tabcolsep}{3.2pt}
\begin{tabular}{llrrl}
\toprule
\textbf{Method} & \textbf{Extraction} & \textbf{Calls} & \textbf{Tokens} & \textbf{Cost source} \\
\midrule
HyperGraphRAG & Facts+entities & 11,656 & 20.9M & Normalized est.\\
PropRAG       & Entities+props. & 23,312 & 21.1M & Reported measure\\
HGRAG         & Entities        & 11,656 &  5.7M & Normalized est.\\
HyperProve    & Entities+facts  & 23,312 & 20.8M & Measured artifacts\\
\bottomrule
\end{tabular}
\end{minipage}
\hfill
\begin{minipage}[t]{0.41\textwidth}
\centering
\textbf{(b) Online inference}\par\smallskip
\scriptsize
\setlength{\tabcolsep}{4.1pt}
\begin{tabular}{lrrr}
\toprule
\textbf{Method} & \textbf{Acc.} & \textbf{Tokens} & \textbf{Lat. (s)} \\
\midrule
BGE-M3     & 51.50 &  1.7M &  4.1K\\
IRCoT      & 56.67 &  4.4M &  3.8K\\
T2RAG      & 54.67 &  2.4M &  9.6K\\
HopRAG     & 67.18 & 36.1M & 45.2K\\
ChainRAG   & 60.01 & 15.4M & 37.0K\\
HyperProve & 71.60 &  9.6M & 31.9K\\
\bottomrule
\end{tabular}
\end{minipage}
\caption{Offline and online efficiency on MuSiQue. Panel (a) explicitly separates measured or reported costs from estimates normalized to the 11,656-passage corpus and the same token accounting. Panel (b) reports measured inference over 1,000 questions. All token and latency values exclude answer-judge cost.}
\label{tab:efficiency_analysis}
\end{table*}

\paragraph{Offline construction.}
HyperProve's measured index uses 23,312 calls and 20.8M tokens, close to PropRAG's reported 23,312 calls and 21.1M tokens and HyperGraphRAG's normalized 20.9M-token estimate; entity-only HGRAG is cheaper at an estimated 5.7M tokens. Passage-local extraction can be parallelized and amortized over queries, although these workloads reflect different extraction objectives. Overall, HyperProve has comparable offline cost to PropRAG and HyperGraphRAG, while the online results below show that this one-time investment yields a favorable accuracy--efficiency trade-off.

\paragraph{Online inference.}
HyperProve uses 9.6M online tokens and 31.9K seconds for 1,000 MuSiQue questions. It improves accuracy over HopRAG by 4.42 points while using 26.5M fewer tokens and 13.3K fewer seconds. Against ChainRAG, it gains 11.59 points while using 5.8M fewer tokens and 5.1K fewer seconds. The lighter BGE-M3, IRCoT, and T2RAG systems remain substantially cheaper, but trail HyperProve by 20.10, 14.93, and 16.93 accuracy points, respectively. Thus, HyperProve occupies a middle accuracy--token--latency regime: more computation than lightweight retrieval, but bounded below the two strongest high-cost iterative baselines evaluated here.

\subsection{Qualitative Analysis}
\label{sec:analysis_qualitative}

Table~\ref{tab:case_multi_entity} presents a multi-entity MuSiQue case with two initially separate evidence chains. One chain resolves \textit{Pathology} to \textit{San Diego}; the other resolves \textit{Still Swingin'} to \textit{Papa Roach} and then connects the band to \textit{Vacaville, California}. Answering the question requires more than recovering these entities: the system must join the two chains and apply the requested rank to the \emph{San Diego urban area within California}, rather than to San Diego as a city.

HyperProve retrieves all four gold passages. At each hop, query rewriting introduces the resolved bridge entity, while the carried hyperedge frontier keeps that entity attached to its supporting fact and source. The first branch therefore preserves the semantic target \textit{San Diego urban area}; the second supplies the geographic constraint \textit{California}. When the branches meet, the final retrieval is conditioned on both pieces of state, allowing the reader to select \textit{third-largest}. In this example, query rewriting provides cross-hop reach, whereas structural propagation preserves provenance and prevents the rank constraint from drifting to a related but different entity type.

ChainRAG covers only two of the four gold passages. It recovers \textit{Pathology} $\rightarrow$ \textit{San Diego} and \textit{Still Swingin'} $\rightarrow$ \textit{Papa Roach}, but misses both the California-binding fact and the evidence that ranks the San Diego urban area. Its answer \textit{2nd} is thus not merely caused by fewer retrieved passages: the incomplete chain leaves the reader with a nearby city-population statistic whose entity scope does not match the question. The comparison illustrates why multi-hop retrieval must preserve relation type, modifiers, and provenance in addition to entity names. Although this single case is not evidence of aggregate superiority by itself, it provides a mechanism-level explanation for the evidence-coverage gains in Section~\ref{sec:analysis_faithfulness}.

\section{Conclusion}
We presented HyperProve, an answer-guided hypergraph expansion framework. HyperProve decomposes complex questions into dependent sub-queries, carries intermediate answers and supporting hyperedges as retrieval state, and performs query-biased local expansion over atomic-fact hyperedges that preserve multi-entity context and source provenance. Across three multi-hop QA benchmarks, HyperProve achieves the strongest performance among dense, structured, and iterative retrieval baselines. These results show the value of combining stateful query control with structured evidence accumulation.

\section*{Limitations}
HyperProve has three limitations. First, it incurs substantial LLM cost during offline extraction and online reasoning; future work could distill these components using recent trajectory, representation, or cross-tokenizer alignment methods~\citep{chi2026mta,dao2026sra,vu2026dwa,truong2026ctpd}. Second, LLM-as-a-judge accuracy may miss subtle reasoning, evidence-sufficiency, and faithfulness errors, motivating human evaluation and stronger evidence-chain metrics. Finally, our experiments are limited to fixed open-domain QA benchmarks with stable corpora and predefined answers, which does not fully capture noisy, incomplete, conflicting, multilingual, or temporally changing evidence.

\section*{Acknowledgments}
This research was funded by Vingroup Innovation Foundation (VINIF) under project code VINIF.2026.DA117.

\bibliography{custom}

\appendix
\appendix

\section{Appendix}

This appendix provides additional details for HyperProve, including the prompts
used for offline hypergraph construction and online answer-guided reasoning,
scalability limitations, dataset statistics, positioning relative to prior work,
and two case studies.

\subsection{Prompt Templates}
\label{app:prompt-templates}

This section lists the prompt templates used by HyperProve for offline
hypergraph construction, online answer-guided retrieval, answer generation,
and automatic evaluation. All prompts are designed to produce structured
outputs, which simplifies downstream parsing and makes the reasoning pipeline
more reproducible.

\subsubsection{Entity Extraction}
\label{app:prompt-entity-extraction}

\begin{promptbox}
Task:
Given a passage, identify the entities that are needed to represent the factual content of the passage.

Output format:
Return only a valid JSON object with the following structure:
{
  "entities": ["entity_1", "entity_2", "..."]
}

Guidelines:
1. Include all named entities mentioned in the passage.
2. Include all dates, temporal expressions, and events.
3. Include generic concepts when they are central to the passage topic.
4. Include entities that participate in meaningful relations with the extracted entities.
5. Preserve the surface form used in the passage whenever possible.
6. Do not return an empty list.
7. Do not include explanations outside the JSON object.

Example passage:
Radio City is India's first private FM radio station and was started on
3 July 2001. It plays Hindi, English and regional songs. Radio City recently
forayed into New Media in May 2008 with the launch of a music portal,
PlanetRadiocity.com, that offers music-related news, videos, songs, and other
features.

Example output:
{
  "entities": [
    "Radio City",
    "India",
    "private FM radio station",
    "3 July 2001",
    "Hindi",
    "English",
    "regional songs",
    "New Media",
    "May 2008",
    "PlanetRadiocity.com",
    "music portal",
    "music-related news",
    "videos",
    "songs"
  ]
}

Input:
Passage: {passage}
\end{promptbox}

\subsubsection{Atomic-Fact Hyperedge Extraction}
\label{app:prompt-hyperedge-extraction}

\begin{promptbox}
Task:
Given a passage and a predefined list of entities, decompose the passage into atomic facts. Each atomic fact should express one self-contained claim and explicitly specify the entities involved in that fact.

Output format:
Return only a valid JSON object with the following structure:
{
  "hyperedges": [
    {
      "text": "a complete atomic fact",
      "entities": ["entity_1", "entity_2", "..."]
    }
  ]
}

Guidelines:
1. Each hyperedge must contain exactly one factual claim or relation.
2. Each hyperedge must be understandable without requiring the original passage.
3. Use only entities from the provided entity list. Do not introduce new entities.
4. Make the relation among entities explicit and unambiguous.
5. Preserve temporal, causal, comparative, and modifier information when it affects the meaning of the fact.
6. When a statement involves more than two entities, keep them together in a single hyperedge rather than splitting the fact into disconnected binary relations.
7. Include all relevant entities from the provided entity list in both the hyperedge text and the entity array.
8. Ensure that the set of hyperedges covers all important factual information in the passage.
9. Do not include explanations outside the JSON object.

Example passage:
In 2020, after Apple launched the M1 chip, major software companies like Adobe optimized their applications, improving performance by up to 80 percent compared to Intel-based Macs.

Entity list:
[
  "Apple",
  "M1 chip",
  "2020",
  "Adobe",
  "Adobe's applications",
  "Intel-based Macs",
  "80% performance improvement"
]

Example output:
{
  "hyperedges": [
    {
      "text": "Apple launched the M1 chip in 2020.",
      "entities": ["Apple", "M1 chip", "2020"]
    },
    {
      "text": "Adobe optimized Adobe's applications for the M1 chip after the chip was launched.",
      "entities": ["Adobe", "Adobe's applications", "M1 chip"]
    },
    {
      "text": "Adobe's applications on the M1 chip achieved up to an 80% performance improvement compared with Adobe's applications on Intel-based Macs.",
      "entities": [
        "Adobe",
        "Adobe's applications",
        "M1 chip",
        "80% performance improvement",
        "Intel-based Macs"
      ]
    }
  ]
}

Input:
Passage: {passage}
Entity list: {entities_json_list}
\end{promptbox}

\subsubsection{Question Decomposition}
\label{app:prompt-question-decomposition}

\begin{promptbox}
Task:
Break the input question into the smallest number of retrieval-friendly sub-questions needed to answer it reliably.

Main principle:
Decompose the question according to bridge facts, not according to surface length. Each sub-question should normally resolve one unknown bridge entity, one bridge relation, or the final requested attribute.

Guidelines:
1. If the question can be answered with a single retrieval step, return a list containing only the original question.
2. For sequential reasoning, create one sub-question for each bridge step when a later step depends on the answer to an earlier step. 
3. A good sub-question should usually do one of the following:
   - identify a bridge entity;
   - identify a relation involving a known bridge entity;
   - ask for the final attribute once the target entity is known.
4. Avoid sub-questions that require resolving multiple new bridge entities at the same time.
5. Prefer two sub-questions for one-bridge questions, three for two-bridge questions, and four for three-bridge questions when needed.
6. Do not force every question into two sub-questions. Use more steps when this makes retrieval clearer.
7. Preserve the reasoning order. Later sub-questions may refer to earlier answers using phrases such as "that person", "that city", or "that work".
8. For parallel or comparative questions, retrieve the required attributes separately before asking for the comparison.
9. Avoid redundant or overlapping sub-questions.
10. Return at most four sub-questions.

Output format:
Return only a valid JSON object:
{
  "sub_questions": [
    "first sub-question",
    "second sub-question"
  ]
}

Examples:
Question:
What county is Erik Hort's birthplace a part of?

Output:
{
  "sub_questions": [
    "What is Erik Hort's birthplace?",
    "Which county is that birthplace part of?"
  ]
}

Question:
When was the start of the battle of the birthplace of the performer of III?

Output:
{
  "sub_questions": [
    "Who is the performer of III?",
    "What is that performer's birthplace?",
    "When did the battle at that birthplace start?"
  ]
}

Input:
Question: {question}
\end{promptbox}

\subsubsection{Sub-query Evaluation}
\label{app:prompt-subquery-evaluation}

\begin{promptbox}
Task:
Given a sub-query, retrieved atomic-fact hyperedges, and source passages, decide whether the evidence answers the sub-query.

Output format:
Return only a valid JSON object:
{
  "status": "ANSWERED" or "UNANSWERED",
  "answer": "short sub-answer or null",
  "supporting_hyperedge_ids": [0, 1]
}

Guidelines:
1. Use ANSWERED only when the hyperedges and passages support a direct answer.
2. The answer must be concise.
3. supporting_hyperedge_ids are 0-based indices from the listed hyperedges.
4. Do not include explanations outside the JSON object.

Sub-query: {sub_query}

Hyperedges:
{hyperedges_block}

Passages:
{passages_block}
\end{promptbox}

\subsubsection{Final Answer Integration}
\label{app:prompt-final-integration}

\begin{promptbox}
Task:
Given the original question and the sub-question answers produced during multi hop reasoning, combine the intermediate answers into a single final answer.

Guidelines:
1. The final answer must directly answer the original question.
2. Use the available sub-question answers.
3. If one sub-answer is missing, still provide the best short answer supported by the available information.
4. Output only the answer, without reasoning or restating the question.
5. Return only a valid JSON object.

Output format:
{
  "final_answer": "..."
}

Original question:
{original_question}

Sub-questions and answers:
{qa_block}
\end{promptbox}

\subsubsection{QA Reader Fallback}
\label{app:prompt-qa-fallback}
\begin{promptbox}
System prompt:
As an advanced reading comprehension assistant, your task is to analyze text passages and corresponding questions meticulously. Your response start after "Thought: ", where you will methodically break down the reasoning process, illustrating how you arrive at conclusions. Conclude with exactly one line formatted as "FINAL_ANSWER: <answer>". The final answer must be a concise answer span only, with no explanation, no restated question, and no extra text after it.

User prompt:
Wikipedia Title: {title_1}
{passage_text_1}

Wikipedia Title: {title_2}
{passage_text_2}

...

Retrieved propositions:
{numbered_propositions}

Retriever hypothesis: {retriever_hypothesis}

Question: {question}
Thought:
\end{promptbox}

\subsubsection{Answer Judge}
\label{app:prompt-answer-judge}

\begin{promptbox}
Task:
Evaluate whether a model prediction correctly answers the question by comparing it against the reference answer.

Decision rule: 
Return [[yes]] if the prediction is factually equivalent to the reference answer or contains the reference answer. Return [[no]] otherwise.

Guidelines:
1. Accept paraphrases, aliases, abbreviations, and equivalent date formats.
2. Accept a response that contains the required answer even if it includes additional reasoning.
3. Return [[no]] if the response contains only part of the required answer.
4. Return [[no]] if the response is broader, narrower, or contradictory.
5. Return only [[yes]] or [[no]].

Question:
{question}

Reference answer:
{ground_truth}

Model prediction:
{prediction}

Is the model prediction correct?
\end{promptbox}

\setcounter{table}{8}
\begin{table*}[t]
\centering
\footnotesize
\setlength{\tabcolsep}{5pt}
\resizebox{\textwidth}{!}{%
\begin{tabular}{@{}lllll@{}}
\toprule
\textbf{Method} & \textbf{Category} & \textbf{Retrieval Unit} & \textbf{Multi-step} & \textbf{Main Role} \\
\midrule
Contriever & Dense Retrieval & Passage & No & Unsupervised dense retrieval baseline \\
BGE-M3 & Dense Retrieval & Passage & No & Strong dense retriever and HyperProve seed retriever \\
HippoRAG2 & Structured RAG & Passage + Triples & Yes & Associative graph retrieval baseline \\
PropRAG & Structured RAG & Propositions & Yes & Structured fact retrieval baseline \\
HyperGraphRAG & Structured RAG & Propositions & Yes & Hypergraph-structured retrieval baseline \\
HGRAG & Structured RAG & Propositions & Yes & Entity--passage hypergraph baseline \\
IRCoT & Iterative Retrieval & Passage & Yes & Retrieval interleaved with reasoning \\
T2RAG & Iterative Retrieval & Triple & Yes & Triplet-driven retrieval and reasoning \\
HopRAG & Iterative Retrieval & Passage & Yes & Retrieve--reason--prune baseline \\
ChainRAG & Iterative Retrieval & Passage + Sentence & Yes & Progressive retrieval baseline \\
HyperProve & Ours & Propositions & Yes & Answer-guided hypergraph expansion \\
\bottomrule
\end{tabular}%
}
\caption{Summary of baseline methods and retrieval units. HyperProve differs from prior methods by combining atomic-fact hyperedges with answer-guided, stateful expansion over local hypergraph neighborhoods.}
\label{tab:baseline-details}
\end{table*}

\setcounter{table}{5}

\subsection{Reader-Fallback Analysis}
\label{app:fallback-analysis}

\begin{table}[t]
\centering
\small
\begin{tabular}{lrrr}
\toprule
\textbf{Depth} & \textbf{Fallbacks} & \textbf{Questions} & \textbf{Rate} \\
\midrule
2-hop & 41 & 518 & 7.92\% \\
3-hop & 38 & 316 & 12.03\% \\
4-hop & 44 & 166 & 26.51\% \\
\midrule
Overall & 123 & 1,000 & 12.30\% \\
\bottomrule
\end{tabular}
\caption{Reader-fallback frequency by annotated reasoning depth on MuSiQue.}
\label{tab:fallback-frequency}
\end{table}

Reader fallback is triggered only when at least one decomposed sub-query remains \textsc{Unanswered}. Its frequency rises from $41/518=7.92\%$ on 2-hop questions to $44/166=26.51\%$ on 4-hop questions. The monotonic increase is consistent with a larger opportunity for an unresolved intermediate step; it does not by itself identify which component caused the failure. Across all depths, the fallback is used for $123/1{,}000=12.3\%$ of questions.

\subsection{Hyperparameter Sensitivity}
\label{app:sensitivity}

\begin{table}[t]
\centering
\small
\setlength{\tabcolsep}{4.2pt}
\begin{tabular}{ccrrrr}
\toprule
$\alpha$ & $\theta$ & \textbf{Recall} & \textbf{EM} & \textbf{F1} & \textbf{Acc.} \\
\midrule
0.50 & 0.40 & 82.31 & 42.83 & 59.23 & 71.60 \\
0.25 & 0.40 & 81.18 & 42.41 & 58.76 & 70.80 \\
0.75 & 0.40 & 83.06 & 43.12 & 59.61 & 71.40 \\
0.50 & 0.30 & 83.45 & 41.97 & 58.49 & 70.30 \\
0.50 & 0.50 & 81.72 & 43.05 & 59.40 & 71.90 \\
\bottomrule
\end{tabular}
\caption{MuSiQue sensitivity to structural-transition weight $\alpha$ and query-transition threshold $\theta$. The first row is the default configuration.}
\label{tab:hyperparameter-sensitivity}
\end{table}

Relative to the default, accuracy changes by at most 1.3 points across the five tested settings, indicating that performance is not sharply sensitive within this local range. Changing $\alpha$ mainly shifts the balance between structural and query-conditioned transitions. Increasing $\alpha$ from 0.50 to 0.75 raises recall from 82.31 to 83.06 and F1 from 59.23 to 59.61, but slightly lowers accuracy from 71.60 to 71.40; decreasing it to 0.25 reduces both recall and accuracy to 81.18 and 70.80. The threshold $\theta$ shows a clearer recall--selectivity trade-off. Lowering it to 0.30 admits more query-relevant neighbors and gives the highest recall, 83.45, but reduces EM, F1, and accuracy. Raising it to 0.50 filters more aggressively, lowering recall to 81.72 while producing the highest accuracy, 71.90. Because no alternative improves retrieval and answer metrics consistently, we retain the balanced default $(\alpha,\theta)=(0.50,0.40)$ rather than selecting a configuration post hoc.

\begin{table}[t]
\centering
\small
\resizebox{\columnwidth}{!}{
\begin{tabular}{lrrrr}
\toprule
\textbf{Dataset} & \textbf{Questions} & \textbf{Passages} & \textbf{Entities} & \textbf{Hyperedges} \\
\midrule
2WikiMultiHopQA & 1,000 & 6,119  & 57,906  & 55,032 \\
HotpotQA        & 1,000 & 9,812  & 112,308 & 99,394 \\
MuSiQue         & 1,000 & 11,656 & 117,894 & 105,542 \\
\bottomrule
\end{tabular}
}
\caption{Dataset statistics for the 1,000-question evaluation subsets.}
\label{tab:dataset-statistics}
\end{table}

\subsection{Scalability Limits}
\label{app:scalability-limits}

The current implementation performs synonym linking as batched all-pairs comparisons and rebuilds global graph artifacts after corpus changes. We have not benchmarked web-scale corpora or dynamic index maintenance, so Table~\ref{tab:efficiency_analysis} should not be read as evidence for those settings. Approximate-nearest-neighbor synonym search and incremental updates are promising future implementation directions, but are not measured capabilities of the present system. Complementarily, recent embedding-model distillation methods such as EMO, SAMD, and TALAS could compress the encoder used for dense seeding and synonym matching~\citep{truong2025emo,tran2026samd,dao2026talas}; this direction targets encoder-side retrieval cost rather than LLM-call cost and has not been evaluated with HyperProve.

\subsection{Dataset Statistics}
\label{app:dataset-statistics}

We evaluate HyperProve on three open-domain multi-hop question answering benchmarks: MuSiQue~\citep{trivedi2022musique}, HotpotQA~\citep{yang2018hotpotqa}, and 2WikiMultiHopQA~\citep{ho2020constructing}. Following PropRAG~\citep{wang2025proprag}, we use the corresponding testing set for each dataset to ensure a consistent and comparable evaluation protocol. For the reported experiments, each benchmark is evaluated on a 1,000-question subset from its testing set, and the corpus passages associated with that split are processed offline to construct HyperProve's atomic-fact hypergraph.

\paragraph{MuSiQue.}
MuSiQue is a multi-hop question answering benchmark designed to test compositional reasoning over multiple supporting passages. Questions in MuSiQue are constructed to reduce shortcut reasoning and require models to combine evidence from distinct reasoning steps before producing the final answer.

\paragraph{HotpotQA.}
HotpotQA is a widely used multi-hop question answering benchmark built from Wikipedia. It contains questions that require reasoning over multiple documents, including bridge-style questions, where an intermediate entity must be identified, and comparison questions, where information about multiple entities must be compared.

\paragraph{2WikiMultiHopQA.}
2WikiMultiHopQA is an entity-centric multi-hop question answering benchmark constructed using Wikipedia and Wikidata. It emphasizes reasoning over structured entity relations and typically requires combining evidence across multiple linked facts or documents.

Table~\ref{tab:dataset-statistics} summarizes the three datasets used in our experiments. These benchmarks cover complementary forms of multi-hop reasoning, including compositional reasoning, bridge reasoning, comparison reasoning, and entity-relation reasoning. In HyperProve, each extracted atomic fact is represented as one hyperedge connected to its participating entities and source passage.

\subsection{Positioning Relative to Prior Work}
\label{app:positioning}

Table~\ref{tab:baseline-details} positions HyperProve relative to the evaluated
baselines by comparing their retrieval units and multi-step control mechanisms.
Dense retrievers rank passages in a single step, which makes them useful for seed
retrieval but leaves them without an explicit cross-hop evidence state.
Structured RAG systems add graph, fact-path, or hypergraph connectivity, but
their expansion is primarily driven by static structure or pre-defined retrieval
paths. Iterative systems introduce multi-step control, although their state is
mainly textual or tied to passages, triples, or sentence-level evidence.

HyperProve combines these directions by retrieving over atomic-fact hyperedges,
preserving $n$-ary entity context and source-passage provenance, and using
intermediate answers to define the next local expansion frontier. This makes
retrieval stateful, fact-centered, and structurally constrained while still
allowing fresh dense seeds to recover evidence outside the carried frontier.

\subsection{Case Study 1: Hidden-Bridge Reasoning}
\label{sec:appendix_case_hidden_bridge}

\begin{figure*}[p]
\centering
\includegraphics[width=\textwidth]{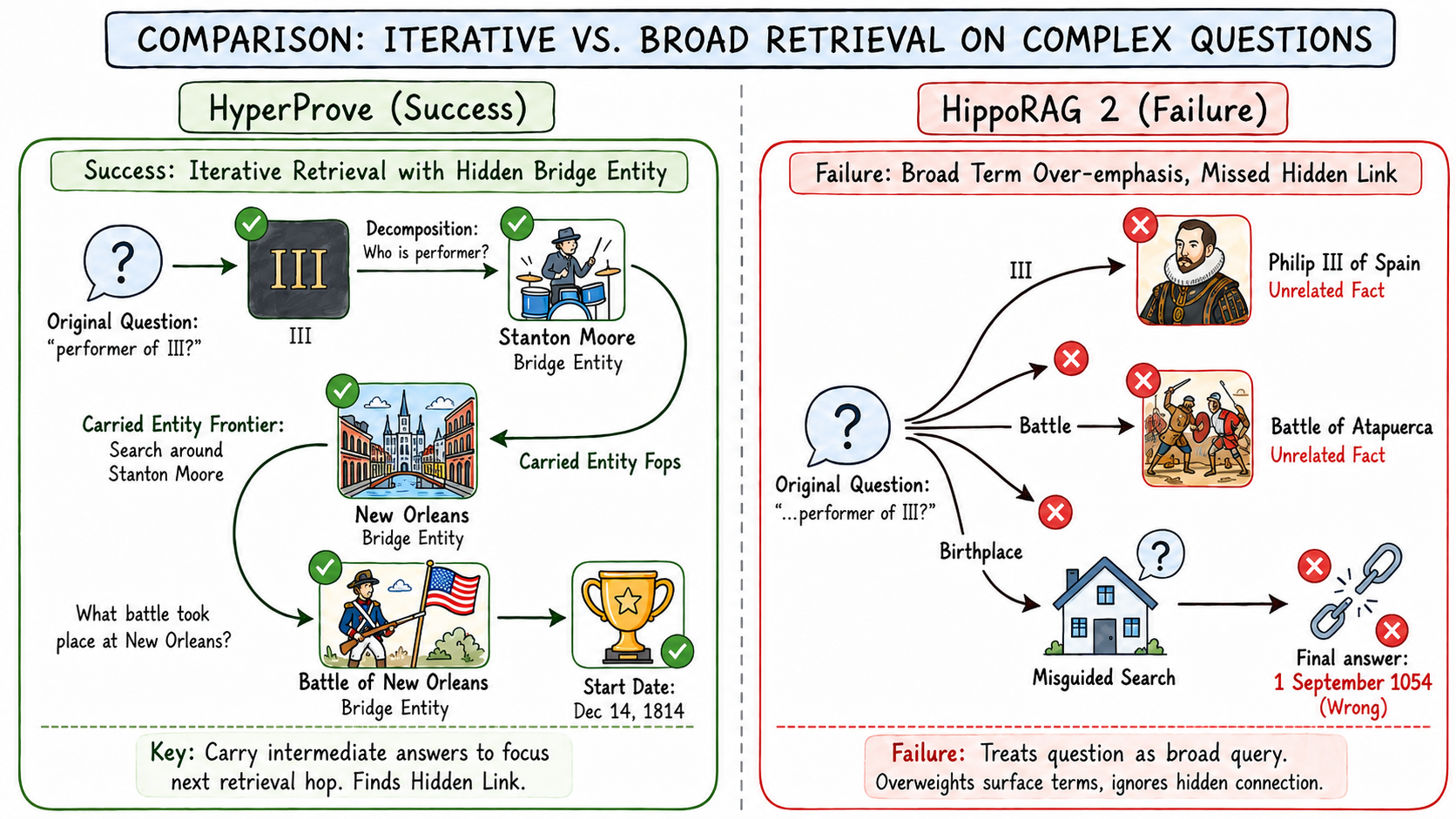}
\caption{Qualitative comparison on a hidden-bridge MuSiQue example. HyperProve carries intermediate answers as bridge entities, enabling retrieval around \textit{Stanton Moore}, \textit{New Orleans}, and the \textit{Battle of New Orleans}. In contrast, HippoRAG2 over-emphasizes broad surface terms from the original question and misses the hidden evidence chain.}
\label{fig:case_hidden_bridge}
\end{figure*}

Figure~\ref{fig:case_hidden_bridge} shows a hidden-bridge MuSiQue example where the question asks: ``When was the start of the battle of the birthplace of the performer of III?'' The original question does not explicitly mention the required bridge entities: \textit{Stanton Moore}, \textit{New Orleans}, or the \textit{Battle of New Orleans}. This makes the example difficult for retrieval methods that mainly rank evidence according to surface similarity with the original question.

HyperProve decomposes the question into dependent sub-queries and carries intermediate answers forward as bridge entities. It first identifies \textit{III} as Stanton Moore's third studio solo album, then uses \textit{Stanton Moore} to retrieve evidence about his birthplace, \textit{New Orleans}. The system then uses \textit{New Orleans} as the next bridge entity to retrieve evidence about the \textit{Battle of New Orleans} and its start date. This answer-guided evidence chain leads to the correct answer, \textit{December 14, 1814}.

HippoRAG2, by contrast, over-emphasizes broad surface terms from the original question, such as \textit{III}, \textit{battle}, and \textit{birthplace}. Because the key intermediate entities are not present in the original question, its associative retrieval process follows unrelated evidence and fails to recover the hidden bridge chain. It therefore produces the incorrect answer, \textit{1 September 1054}. This case highlights the importance of answer-conditioned bridge propagation: in multi-hop QA, the next useful evidence is often not lexically similar to the original question, but becomes reachable only after an intermediate entity has been resolved.

\subsection{Case Study 2: Cascading Answer State}
\label{app:failure-case}

The gold chain in this four-hop example is \textit{California} $\rightarrow$ \textit{Samuel Brannan} $\rightarrow$ \textit{Sacramento} $\rightarrow$ \textit{Rio Linda}, whereas HyperProve predicts \textit{California} $\rightarrow$ \textit{Francis J. Banfield} $\rightarrow$ \textit{New York City} $\rightarrow$ \textit{New Jersey}. The first hop resolves the location correctly, but the second accepts Francis J. Banfield as a plausible yet unsupported person. Because this answer conditions both the rewritten query and the structural frontier, the third hop follows evidence toward New York City rather than Sacramento. The final hop then retrieves locally coherent evidence around the wrong frontier and returns New Jersey instead of Rio Linda.

This case exposes the main risk of answer-guided retrieval: later evidence can remain coherent with a propagated state while becoming inconsistent with the original question. Confidence-aware propagation could defer uncertain bridges; provenance constraints could reject unsupported frontiers; and multi-frontier recovery could retain alternatives. These remain unevaluated future directions.

\clearpage
\clearpage

\end{document}